\documentclass[12pt]{article}
\usepackage{times}  
\usepackage{helvet}  
\usepackage{courier}  
\usepackage[hyphens]{url}  
\usepackage{graphicx} 
\usepackage{caption} 
\usepackage{algorithm}
\usepackage{algorithmic}
\usepackage{subcaption} 
\usepackage{bm}
\usepackage{amsmath}
\usepackage{xcolor}
\usepackage{color,soul}

\usepackage{newfloat}
\usepackage{listings}
\DeclareCaptionStyle{ruled}{labelfont=normalfont,labelsep=colon,strut=off} 
\floatstyle{ruled}
\newfloat{listing}{tb}{lst}{}
\floatname{listing}{Listing}
\usepackage[linktoc=all]{hyperref}
\hypersetup{colorlinks=true,
	linkcolor=blue,
	citecolor=blue
}

\usepackage[margin=1in]{geometry}
\usepackage{titlesec}

\titleformat{\paragraph}
{\normalfont\normalsize\bfseries}{\theparagraph}{1em}{}
\titlespacing*{\paragraph}
{0pt}{3.25ex plus 1ex minus .2ex}{1.5ex plus .2ex}

\title{Spatio-Temporal Super-Resolution of Dynamical Systems using Physics-Informed Deep-Learning}
\author{
    Rajat Arora\thanks{Senior member of technical staff at Advanced Micro Devices, Inc. (AMD).}, 
    Ankit Shrivastava\thanks{Post Doctoral associate at Sandia National Laboratories.}
}
\newsavebox{\largestimage}

\newcommand\blfootnote[1]{%
  \begingroup
  \renewcommand\thefootnote{}\footnote{#1}%
  \addtocounter{footnote}{-1}%
  \endgroup
}

\begin{document}
	
\pagenumbering{arabic}





\date{}

\maketitle
\begin{abstract}




This work presents a physics-informed deep learning-based super-resolution framework to enhance the spatio-temporal resolution of the solution of time-dependent partial differential equations (PDE). Prior works on deep learning-based super-resolution models have shown promise in accelerating engineering design by reducing the computational expense of traditional numerical schemes. However, these models heavily rely on the availability of high-resolution (HR) labeled data needed during training.

In this work, we propose a physics-informed deep learning-based framework to enhance the spatial and temporal resolution of coarse-scale (both in space and time) PDE solutions without requiring any HR data. The framework consists of two trainable modules independently super-resolving the PDE solution, first in spatial and then in temporal direction. The physics based losses are implemented in a novel way to ensure tight coupling between the spatio-temporally refined  outputs at different times and improve framework accuracy. We analyze the capability of the developed framework by investigating its performance on an elastodynamics problem. It is observed that the proposed framework can successfully super-resolve (both in space and time) the low-resolution PDE solutions while satisfying physics-based constraints and yielding high accuracy. Furthermore, the analysis and obtained speed-up show that the proposed framework is well-suited for integration with traditional numerical methods to reduce computational complexity during engineering design.


\blfootnote{Accepted at AAAI 2023: Workshop on AI to Accelerate Science and Engineering (AI2ASE)}

\end{abstract}


\section{Introduction}
Accurate modeling of the dynamic behavior of nonlinear systems is crucial for many industrial applications ranging from microscale MEMS sensors to large-scale structural systems. Therefore, significant research is being done to understand and resolve the complex physical phenomena occurring within these dynamical systems at extremely small spatial and temporal scales. This scientific pursuit of capturing complex physical phenomena occurring at widely varying spatio-temporal scales has led to the ever-increasing sophistication of the physical system's governing Partial Differential Equations (PDEs). For example, a PDE-based model \cite{arora2020dislocation,arora2020unification, joshi2020equilibrium, arora2020finite,arora2019computational, arora2022mechanics} capturing defects evolution in materials at the nanoscale has been shown superior to conventional theories for a wide range of applications. However, the massive data storage and computational expense requirements to simulate such multi-physics coupled PDEs with high-fidelity bring traditional numerical solvers to their limits. Hence, fast and accurate techniques to perform these multi-physics simulations at multi-scale are of utmost importance.

On the other hand, the recent advances in Machine Learning (ML) have led to the development of several data-driven and Physics Informed ML models to solve PDEs occurring in fluid \cite{sun2020surrogate,rao2020physics,jin2021nsfnets} and solid mechanics \cite{frankel_prediction_2020, arora2022physics, shrivastava2022predicting, zhu2021machine}. However, issues ranging from its theoretical considerations (such as convergence, stability, accuracy, and generalizability) to issues related to boundary conditions, neural network architecture design, or optimization aspects still need to be fully resolved  \cite{markidis2021old, cuomo2022scientific}. Therefore, hybrid strategies integrating physics-informed ML with traditional approaches are emerging as a promising option to tackle this computational challenge of solving complex multi-physics PDEs \cite{arora2021machine, arora2022physrnet, gao2021super}. 

To this end, in this research, we aim to investigate a two-stage hybrid approach integrating ML and traditional approaches to obtain (reconstruct) solutions to spatio-temporal PDEs. 1) In the first stage, low-resolution (LR) PDE solutions are obtained by doing numerical simulations on a coarse scale both in space and time (using large grid size and timestep). This low-resolution solution with satisfactory accuracy can be generated with a huge reduction in computational expense compared to solving PDE on a fine scale. 2) In the second stage, the spatio-temporal resolution of this coarse-scale solution is enhanced using a physics-formed deep learning-based framework. A significant advantage of such `physics-guided resolution enhancement' approach is the reduced computational expense and data storage requirements during the scientific exploration phase, which will significantly accelerate the process of scientific investigation and engineering design. This enhancement in resolution will also be referred to as the upsampling or super-resolution (SR) in this work.



The recent works involving spatio-temporal super-resolution of physical systems \cite{ren2022physics, esmaeilzadeh2020meshfreeflownet, fukami2021machine} use labeled high-resolution (HR) ground truth data for model training. Fukami et.~al \cite{fukami2021machine} presents a purely data-driven SR framework, and therefore the super-resolved fields may not satisfy the physics-based constraints accurately. The works of Ren et al. \cite{ren2022physics} and Soheil et al. \cite{esmaeilzadeh2020meshfreeflownet} are `almost data-driven' in that the scaling coefficient of prediction/data loss is chosen to be $20$ times the coefficient of physics loss in the total loss for optimal accuracy. In fact, the errors are huge when HR-labeled data is not taken into account (purely physics-driven) \cite[see Table 1]{esmaeilzadeh2020meshfreeflownet}. Moreover, the HR-labeled data is computationally expensive to obtain. Therefore, its use during training completely negates the massive benefit of accelerating scientific computing that ML models aim to achieve. In this research,  we present an end-to-end physics-informed deep learning-based framework to enhance the spatial and temporal resolution of coarse scale (both in space and time) PDE solutions without requiring any HR-labeled data. In summary, our main contributions are as follows: 

\begin{enumerate}

\item We propose a novel and efficient physics-informed deep learning-based spatio-temporal resolution enhancement framework.
    
\item The framework consists of two trainable deep learning modules independently responsible for spatial and temporal upscaling of the coarse-scale PDE solution.

\item The physics based losses are implemented in a novel way to ensure tight coupling between the spatio-temporally refined  outputs at different times and improve framework accuracy.

\item Unlike other works \cite{esmaeilzadeh2020meshfreeflownet, fukami2021machine, ren2022physics}, the proposed framework does not rely on the availability of any high-resolution labeled data.
    
\item The effectiveness of the framework is tested by using the low-resolution coarse grid simulation data as input as opposed to using downsampled high-resolution labeled data.

\end{enumerate}

\textbf{Organization.} The remainder of this paper is organized as follows:
Section \ref{sec:background}  briefly recalls the governing equations of a general spatio-temporal PDE. Section \ref{sec:methodology} present the details of the framework architecture, data setup, and composite loss function. Towards the end, Section \ref{sec:results} presents the results that validate the developed framework and demonstrate its effectiveness in super-resolving the solution fields for the test problem discussed. Finally, conclusions and avenues for further research are briefly discussed in Section \ref{sec:conclusion}.

\section{Background}\label{sec:background}
\subsection{Governing equations}
\label{sec:gen_gov_eq}

A typical spatio-temporal PDE governing the dynamical systems can be written in the following form:
\begin{align}
\begin{split}
\dot{\bm{z}} - \bm{\mathcal{F}}(\bm{z}, \bm{x}, t; \bm{x}\in\bm{\Omega}) = \bm{0},
\label{eq:dynamics}
\end{split}
\end{align}
subjected to the initial and boundary conditions 
\begin{align}
\begin{split}
    \bm{\mathcal{I}}(\bm{z};~ t=0, \bm{x}\in\bm\Omega) = \bm{0},\\
    \bm{\mathcal{B}}(\bm{z};~ \bm{x}\in\partial\bm{\Omega}) = \bm{0}.
    \label{eq:dynamics_icbc}
\end{split}
\end{align}
In equations \ref{eq:dynamics}, \ref{eq:dynamics_icbc}, $\bm{z}$ denotes the system solution comprised of $m$ state variables, and $\dot{\bm{z}}$ denotes its time derivative. $\bm{\mathcal{F}}$ is the nonlinear functional of the polynomial and derivatives terms of its arguments. $\bm{\Omega}$ and $\partial\bm\Omega$ denotes the physical domain and its boundary, respectively. Any additional constraints which are inherently present in the system or required because of numerical methods, such as mixed finite element methods, can be assembled into $\bm{\mathcal{C}}(\bm{z}) = \bm{0}$.  

\begin{figure*}[t]
\centering
\begin{subfigure}{.85\textwidth}
  \includegraphics[width=\linewidth]{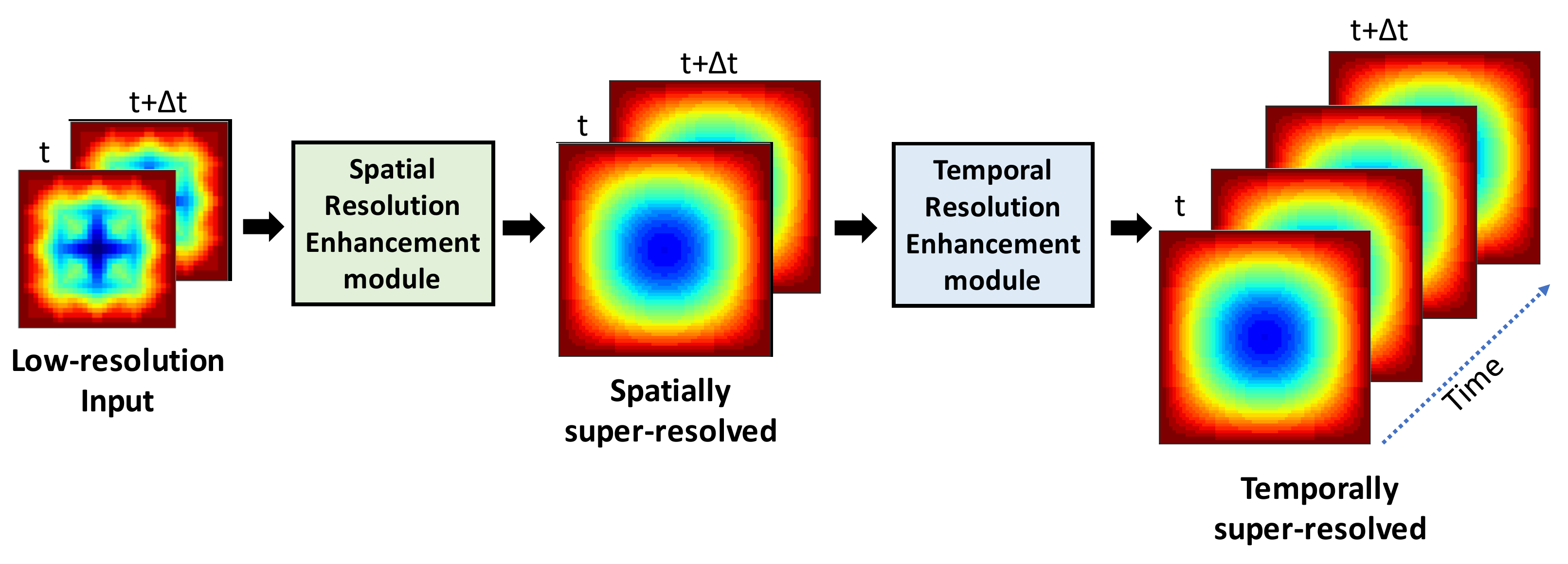}
 \end{subfigure}
  \caption{
  The figure shows the structure of the proposed framework.
  The input to the framework is a two-channel image of the coarse-scale solution, and the final output of the framework is a ($k+1$) channel image of the spatio-temporally resolved solution. The framework consists of a spatial and a temporal resolution enhancement module. The former performs super-resolution in space, whereas the latter super-resolves in the temporal direction. The modules are trained independently.
  }
  \label{fig:RDN}
\end{figure*}


Given a solution $\bm{z}_c$ which is obtained by solving the system of equations \eqref{eq:dynamics}-\eqref{eq:dynamics_icbc} at a coarse scale (large mesh size and timestep), our objective is to enhance the spatial and temporal resolution of the solution by using a physics-informed deep-learning based method. The schematic of our proposed framework is shown in Figure \ref{fig:RDN}.

\section{Methodology}
\label{sec:methodology}
Section \ref{sec:loss} briefly outlines the physics-informed composite loss (objective function) used during the framework's training. We then present an overview of the ``end-to-end" spatio-temporal resolution enhancement framework in Section \ref{sec:model_frame}.


%

%

\subsection{Objective function} \label{sec:loss}
We note that the framework proposed in this work is unsupervised and therefore the composite loss function is obtained only from the governing equations of the system - Initial conditions, boundary conditions, and PDEs. Following \cite[Sec. IV]{arora2022physrnet}, we impose the boundary conditions in the `hard' manner (exactly), thus eliminating the boundary condition loss contribution from the composite loss. The physics-informed objective function is then written as follows:
\begin{align}
	\begin{split}
		\mathcal{L} ~=~ &\lambda_1\, \left( \underbrace{||\dot{\bm{z}} - \bm{\mathcal{F}(\bm{z},\bm{x}, t)}||_1}_{\text{PDE}}\right) \, + \\ & \lambda_2 \,\underbrace{||\,\bm{\mathcal{C}}(\bm{z})||_1}_{\text{Constraints}}   ~~ +  ~~ \lambda_3 \, \underbrace{ ||\,\bm{\mathcal{I}}(\bm{z})||_1  }_{\text{I.Cs.}},
	\end{split}
\label{eq:general_loss}
\end{align} where $||A||_1$ denotes the mean absolute error (MAE) between each element in the quantity $A$ and target $0$. 

We use a fourth-order finite difference scheme to evaluate the spatial derivatives of the solution over the grid. For time discretization, we use the Crank-Nicholson algorithm, which has the virtues of being unconditionally stable and is also second-order accurate in both space and time dimensions. 


\subsection{Input and output for the framework}
The input to the framework consists of tuples of LR coarse-scale PDE solution $\bm{\mathcal{I}}_c^t=\{\bm{z}_c^t,~ \bm{z}_c^{t+\Delta t}\}$ at the consecutive timesteps $t$ and $t+\Delta t$, respectively. The outputs from the framework consist of the spatially upscaled PDE solution at the same timesteps $\{\hat{\bm{z}}^t, \hat{\bm{z}}^{t+\Delta t} \}$ along with the synthesis of the HR snapshots at $(k-1)$ intermediate timesteps $\{\hat{\bm{z}}^{t+\frac{\Delta t}{k}}, \hat{\bm{z}}^{t+2\frac{\Delta t}{k}}, ...,  \hat{\bm{z}}^{t+(k-1)\frac{\Delta t}{k}}\}$. Therefore, the framework produces spatially upscaled PDE solution at $(k+1)$ timesteps referred to  $\bm{\mathcal{O}}^t$. We refer to the scalar $k$  as the temporal upscaling factor and is given as the ratio of the coarse-scale timestep to the fine-scale timestep i.e.~$k=\frac{\Delta t_c}{\Delta t_f}$. Similarly, the upscaling factor in the spatial direction $s$ is given as the ratio of the coarse to fine grid resolution, i.e.~$s = \frac{\Delta x_c}{\Delta x_f}$. 



\begin{figure*}[t]
\centering
\begin{subfigure}{.70\textwidth}
  \includegraphics[width=\linewidth]{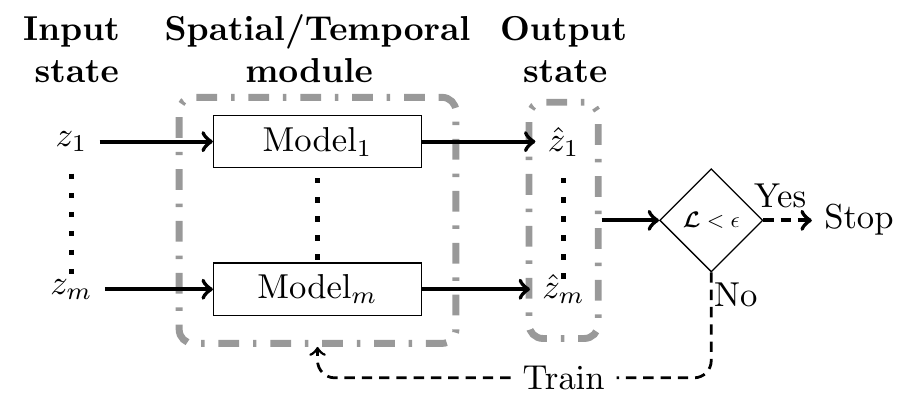}
\end{subfigure}
  \caption{
  The figure shows the structure of the spatial/temporal module.
  Each module consists of $m$ deep-learning models independently super-resolving the state variables $\{z_1, z_2, \dots, z_m\}$.
  The models are, however, trained using a coupled loss function.}
  \label{fig:model_block}
\end{figure*}

\subsection{Framework Architecture} \label{sec:model_frame}
The framework is composed of two trainable modules: the Spatial resolution enhancement module and the Temporal resolution enhancement module, as shown in Figure \ref{fig:RDN}. These two modules independently perform the super-resolution in space and time, respectively. We observed that this dual module approach which first performs super-resolution in space and subsequently increases temporal resolution leads to better convergence and accuracy of the super-resolved fields as is also observed in the data-driven approach of Fukami et.~al \cite{fukami2021machine}. 

A typical solution $\bm{z}$ of any dynamical system consists of $m$ state variables. Therefore, each module in the framework consists of $m$ deep learning models (with the same architecture) that individually reconstruct each state variable; refer to figure \ref{fig:model_block}. These models, however, are coupled during the training through the objective function (loss). Sections \ref{sec:spatial} and \ref{sec:temporal} discuss these spatial and temporal super-resolution modules in greater detail, respectively.




\begin{figure*}[t]
\centering
\begin{subfigure}{.75\textwidth}
  \includegraphics[width=\linewidth]{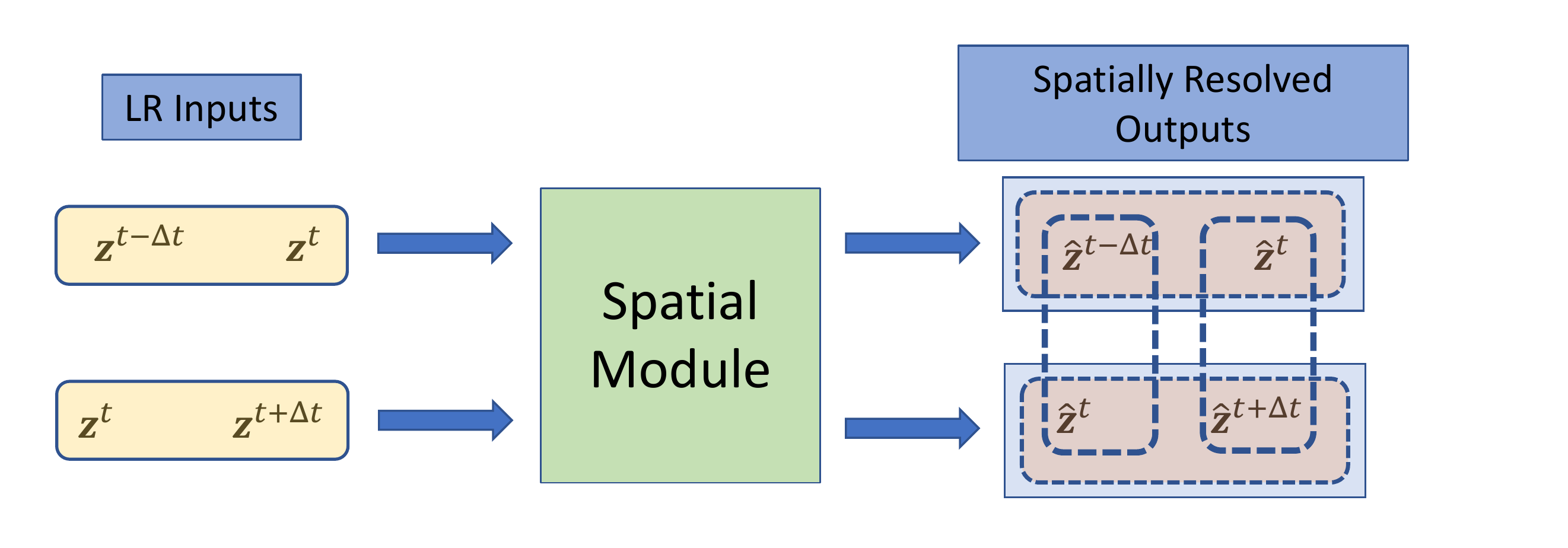}
 \end{subfigure}
  \caption{The figure shows the coupling between the predictions from the spatial resolution enhancement module while calculating the time derivative in the physics-based PDE loss. The two dashed vertical boxes show state variables from different outputs to calculate the time derivatives. Similarly, the two dashed horizontal boxes show the state variables within the same outputs that are used to calculate the time derivative.}
  \label{fig:loss_dia1}
\end{figure*}

\subsubsection{Spatial resolution enhancement module} \label{sec:spatial}
 Given a tuple of LR snapshots at two consecutive time steps $\bm{\mathcal{I}}^t =\{\bm{z}_c^t, \bm{z}_c^{t+\Delta t}\}$ for the state vector $\bm{z}$, the spatial upscaling module outputs the corresponding HR frames $\{\hat{\bm{z}}^t, \hat{\bm{z}}^{t+\Delta t}\}$ representing the enhanced spatial resolution of the input state at the original time steps. Therefore, the input (output) to this module consists of a two-channel image representing the values of the LR (HR) state variables at time steps $t$ and $t+\Delta t$.
 
 During training, we observe that for the successful evolution of solution from initial conditions in both the output channels, the PDE loss in \eqref{eq:general_loss} has to be implemented both within an output and across outputs as highlighted in Figure \ref{fig:loss_dia1}. This coupling in the loss helps in mitigating the \textit{propagation failure} mode \cite{daw2022rethinking} for this module.
 
 Each model in the spatial upscaling module is built upon the Residual Dense Network (RDN) proposed in \cite{zhang2018residual}, which has unique advantages for image SR over other networks \cite[Sec. 4]{zhang2018residual}. We use $4$ residual blocks with $8$ layers in each block and a feature channel size of $32$. The kernel size for convolution is set to be $3$.


\begin{figure*}[t]
\centering
\begin{subfigure}{.75\textwidth}
  \includegraphics[width=\linewidth]{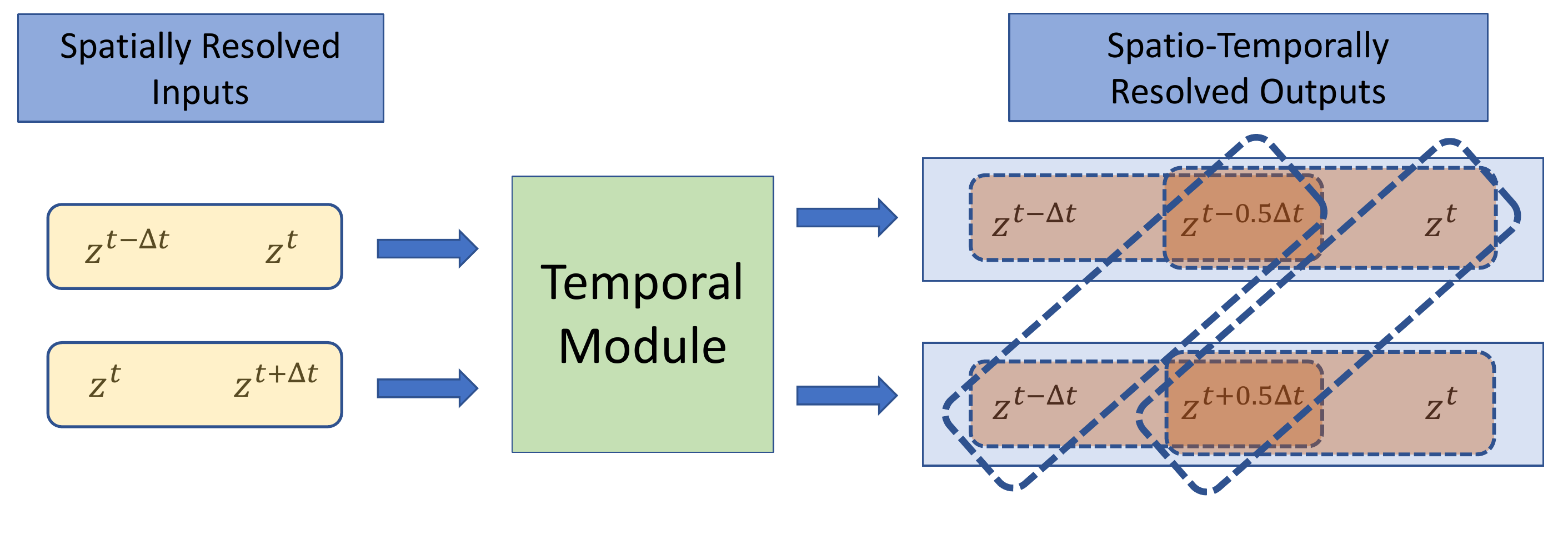}
\end{subfigure}
  \caption{The figure shows the coupling between the predictions from the temporal resolution enhancement module while calculating the time derivative in the physics-based PDE loss. The (two) dashed inclined boxes show state variables from different outputs to calculate the time derivatives. Similarly, the four dashed horizontal boxes show the state variables within the same outputs that are used to calculate the time derivative.}
  \label{fig:loss_dia2}
\end{figure*}

\subsubsection{Temporal resolution enhancement module} \label{sec:temporal}



The temporal module enhances the resolution of the state variables in time. The output of the spatial module serves as an input to the temporal module. Therefore, the input is a two-channel image representing the values of the HR state variables at time steps, $t$, and $t+\Delta t$. The module outputs an image with $(k+1)$ channels, representing the state variables at time steps $\{t$, $t+\frac{\Delta t}{k}$, \dots, $t+\Delta t\}$. We note here that the inputs to the temporal module still have the temporal discretization error of the order O($\Delta t^2$). Therefore, the temporal module also reconstructs the solution at the initial and the final times $t$ and $t+\Delta t$, respectively, to reduce this error to $O\left(\left(\frac{\Delta t}{k}\right)^2\right)$.  We make the following changes to the composite loss \eqref{eq:general_loss} for the training of this module:

\begin{itemize}
   \item Similar to the implementation of PDE loss in the spatial module, the PDE loss for the temporal module is also implemented both within an output and across outputs, as highlighted in Figure \ref{fig:loss_dia2}. 
    \item Since we reconstruct the outputs at initial and final timesteps ($t$ and $t+\Delta t$), we add a constraint loss between these inputs and outputs, which helps in faster convergence of the module.
\end{itemize}

For the sake of simplicity, the model architecture is similar to the spatial module except for the number of channels in the output and the elimination of upsampling layer.

\section{Experiment}
\label{sec:results}
In this section, we briefly discuss the problem setup, dataset generation, and evaluation metric. We then evaluate the performance of the developed framework by investigating its effectiveness in enhancing the spatio-temporal resolution of the coarse-scale solutions to an elastodynamics problem. The results presented herein demonstrate the remarkable accuracy of the network without requiring any HR-labeled data, unlike all previous works.

\subsection{Setup}
\label{sec:setup}
We consider a mixed-variable elastodynamics system widely used in structural engineering and seismologic applications. The governing equations of the system (in the absence of inertia) under $2$-d antiplane strain conditions are given as follows:
\begin{align}
\begin{split}
    v =& ~ \dot{u} \\
    \dfrac{\partial \sigma_{xz}}{\partial x} ~ +&~\dfrac{\partial \sigma_{yz}}{\partial y} = \frac{\rho}{\mu} \dot{v}\\
    \sigma_{xz} = \mu \frac{\partial u} {\partial x} ~~;&~~ \sigma_{yz} = \mu \frac{\partial u} {\partial y}, \label{eq:gov-eq}
\end{split}
\end{align}
where $\rho$ is the material density, $b$ is the body force per unit volume, $\mu$ denotes the shear modulus of the material, and $u$ and $v$ denote the displacement and material velocity (both in the $z$ direction),  respectively. $\sigma_{xz}$ and $\sigma_{yz}$ denote the nonzero components of the (symmetric) stress tensor $\bm{\sigma}$. The equations \ref{eq:gov-eq}, along with the boundary and initial conditions in the equations \ref{eq:bc}, define the governing equations of the system. 
\begin{align}
\begin{split}
{u(t)} = {u}_{bc} &\textrm{~on~} \partial\Omega,\\
{u}(t=0) = {u}_0 ~~ &\textrm{and} ~~ {v}(t=0) = {v}_0.
\end{split}
\label{eq:bc}
\end{align}
In the above, ${u}_0$ and ${v}_0$ denote the known initial conditions on ${u}$ and ${v}$, respectively. ${u}_{bc}$ denote the known displacement on the domain boundary $\partial\Omega$. Without loss of generality, we take ${v}_0 = {u}_{bc} = {0}$ in this work. The initial condition for $u$ is $u_0 = \sin(\pi x)\sin(\pi y)$. 



\subsection{Generation of low-resolution input data}
\label{sec:input_data}
To evaluate the performance of the framework, we generate the low-resolution input data for the problem setup presented above in Section \ref{sec:setup}. We solve the governing system of equations \eqref{eq:gov-eq} for the non-dimensional values of stress, displacement, and velocity, which amounts to setting $\frac{\rho}{\mu} = 1$ in \eqref{eq:gov-eq}. The equations are solved on a coarse mesh of $64$ triangular elements ($41$ nodes and $\Delta x_c\approx0.176$) and coarse-scale timestep $\Delta t_c = 0.005$ with finite element method (FEM) for $48$ timesteps. The obtained coarse grid solution is then interpolated to a structured $32\times32$ grid using the FEM interpolation and is used as an input to the super-resolution framework. For comparison, the HR ground truth data is obtained by solving the same equations on a fine mesh of $64\times64$ nodes and $\Delta x_f=.0158$ with   $\Delta t_f = \frac{\Delta t_c}{2}$. This sets the spatial and temporal upscaling factors to be $s=11$ and $k=2$, respectively, for the test case discussed.


\subsection{Evaluation metric}
For any state variable $\alpha$ (one of $u, v, \sigma_{xz},$ or $\sigma_{yz}$), we define a full field error measure $e$ to quantitatively measure the discrepancy between the HR ground truth data $\alpha^{HR}$ and the framework predictions $\hat{\alpha}$ as follows:
\begin{align}
    e = \dfrac{||\alpha^{HR} - \hat\alpha||_{L^2}} {||\alpha^{HR}||_{L^2}} \times 100.
\end{align}
We note here that the HR ground truth data is only used for comparison with the predicted outputs of the framework.
\subsection{Training}
The framework is implemented and trained using PyTorch. The training strategy used in this work consists of two stages: a) The spatial module is trained in the first stage, which amounts to enhancing spatial resolution. b)  The outputs from the trained spatial module are then used as input to the temporal module during its training. The weights for the spatial module are frozen during the second stage. 

In both stages, we use Adam optimizer with a learning rate of $4\times10^{-4}$ for around $2000$ epochs with a batch size of $8$ samples. As the training progresses, the learning rate is adjusted using  \texttt{ReduceLROnPlateau} scheduler with \texttt{patience} set to $40$. We use a sequential training data sampler to respect the causal structure inherently present in the spatio-temporal PDEs, which has been shown to improve the accuracy of the physics-informed neural networks significantly, refer \cite{wang2022respecting}. We use the scaling coefficients $\lambda_1=5$, $\lambda_2=1$, and $\lambda_3=10$ in this work.


\begin{figure*}[t]
\centering
\begin{subfigure}{.48\textwidth}
  \includegraphics[width=\linewidth]{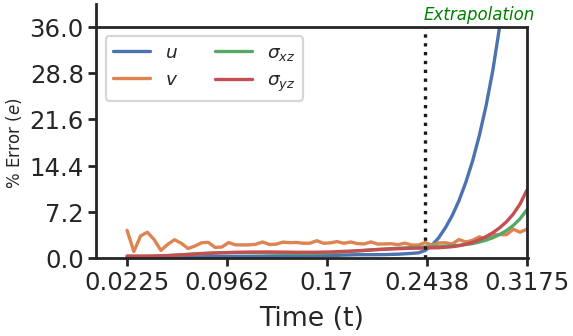}
  \caption{Proposed framework.}
  \label{fig:error_model}
\end{subfigure}%
\hspace*{\fill}
\begin{subfigure}{.48\textwidth}
  \centering
  \includegraphics[width=\linewidth]{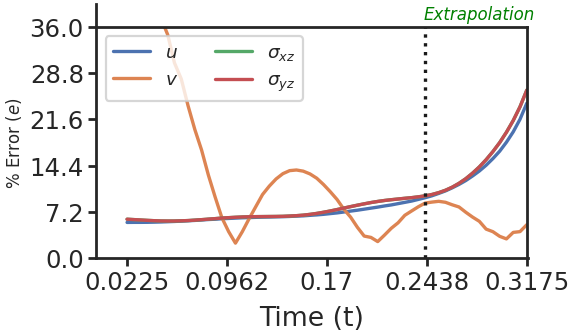}
  \caption{Bi-linear interpolation.}
  \label{fig:error_us}
\end{subfigure}
\caption{
The figure shows the error for the solution fields of elastodynamics at different time steps after super-resolution using the proposed framework, Figure \ref{fig:error_model}, and using bi-linear interpolation, Figure \ref{fig:error_us}.
The super-resolution framework is highly accurate as compared to simple bi-linear interpolation.
The framework's accuracy suffers beyond time, $t=0.2438$ (extrapolation region), as the framework is trained till $t=0.2438$.
The high error in the extrapolation region suggests that the framework is capable of producing high-accuracy results for super-resolution within the convex hull of training set.}
\label{fig:b1}
\end{figure*}

\begin{figure}[t]
	\centering
	\begin{subfigure}[b]{.220\linewidth}
		\centering
	\hspace{10pt}	\small \text{LR input}\vspace{2pt}\par
		{\includegraphics[width=0.995\linewidth]{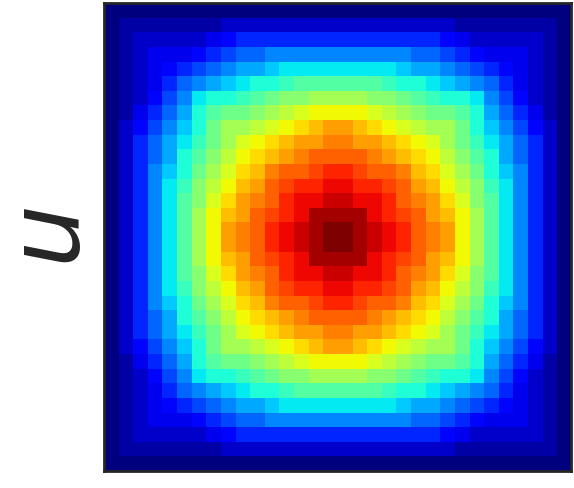}}
	\end{subfigure}%
	\begin{subfigure}[b]{.220\linewidth}
		\centering
		\hspace{14pt} {\footnotesize {HR Ground Truth}}\vspace{4pt}\par
		{\includegraphics[width=0.995\linewidth]{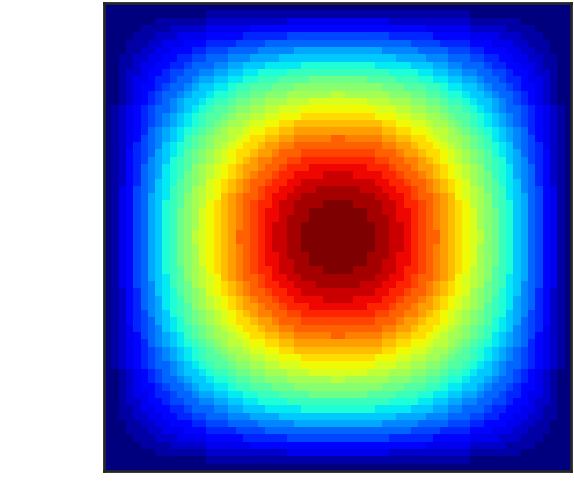}}
	\end{subfigure}%
	\begin{subfigure}[b]{.220\linewidth}
		\centering
	\hspace{14pt}	\small Framework outputs\vspace{2pt}\par
		{\includegraphics[width=0.995\linewidth]{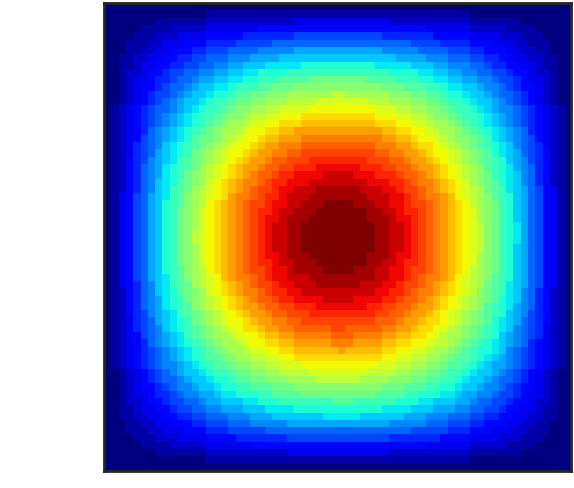}}
	\end{subfigure}%
    \begin{subfigure}[b]{.220\linewidth}
		\centering
	\hspace{8pt}	\footnotesize {Bi-linear interpolation} \vspace{2pt}\par
		{\includegraphics[width=0.995\linewidth]{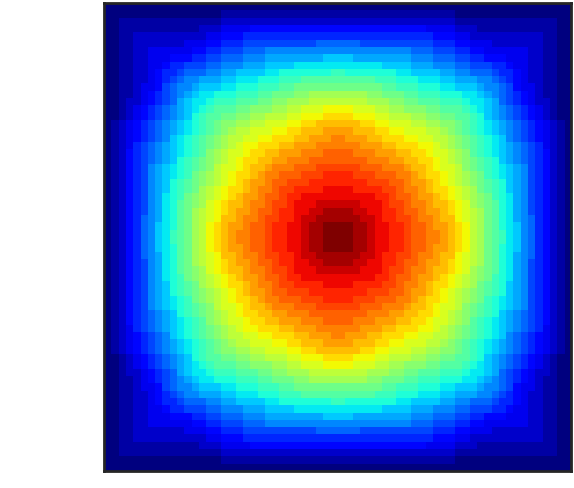}}
	\end{subfigure}\\
	\begin{subfigure}[b]{.220\linewidth}
		\centering
		{\includegraphics[width=0.995\linewidth]{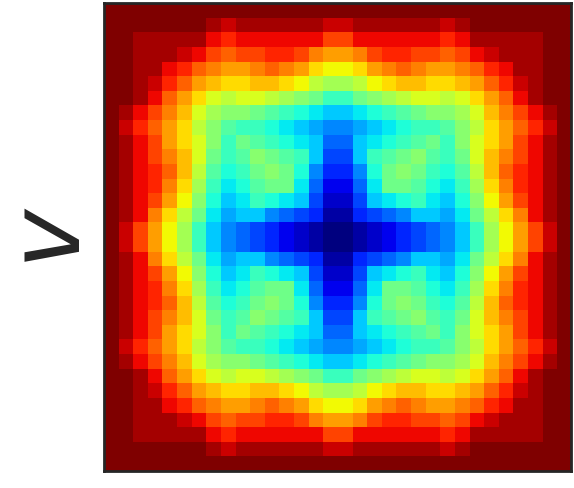}}
	\end{subfigure}%
	\begin{subfigure}[b]{.220\linewidth}
		\centering
		{\includegraphics[width=0.995\linewidth]{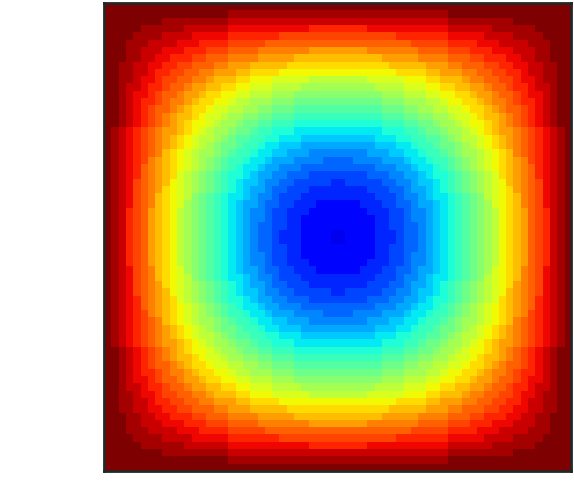}}
	\end{subfigure}%
	\begin{subfigure}[b]{.220\linewidth}
		\centering
		{\includegraphics[width=0.995\linewidth]{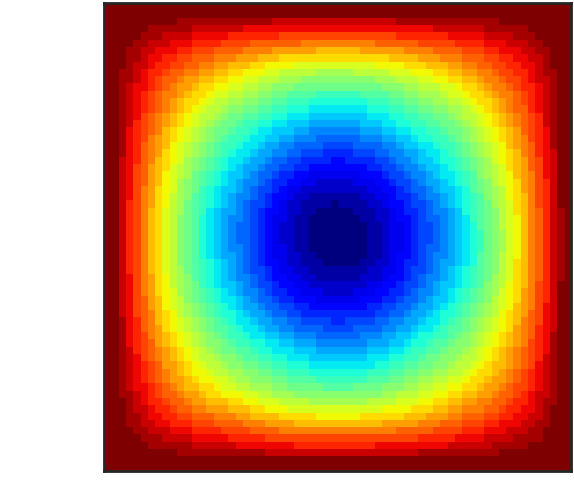}}
	\end{subfigure}%
    \begin{subfigure}[b]{.220\linewidth}
		\centering
		{\includegraphics[width=0.995\linewidth]{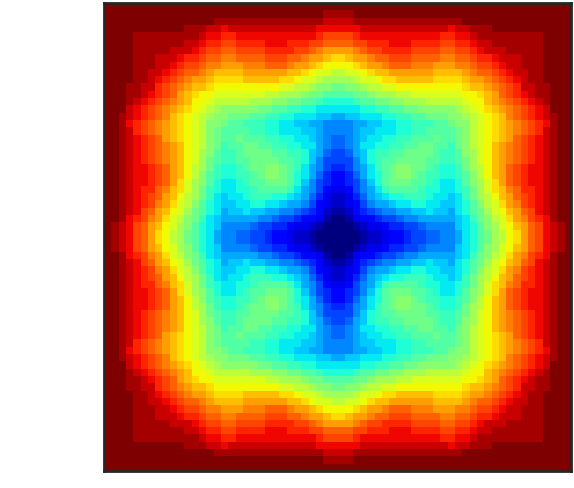}}
	\end{subfigure}\\
	\begin{subfigure}[b]{.220\linewidth}
		\centering
		{\includegraphics[width=0.995\linewidth]{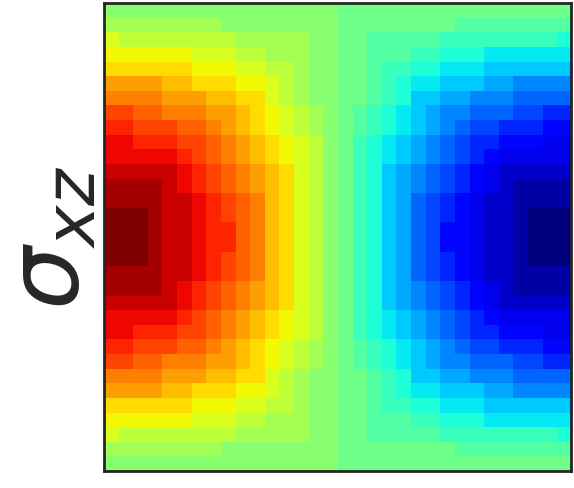}}
	\end{subfigure}%
	\begin{subfigure}[b]{.220\linewidth}
		\centering
		{\includegraphics[width=0.995\linewidth]{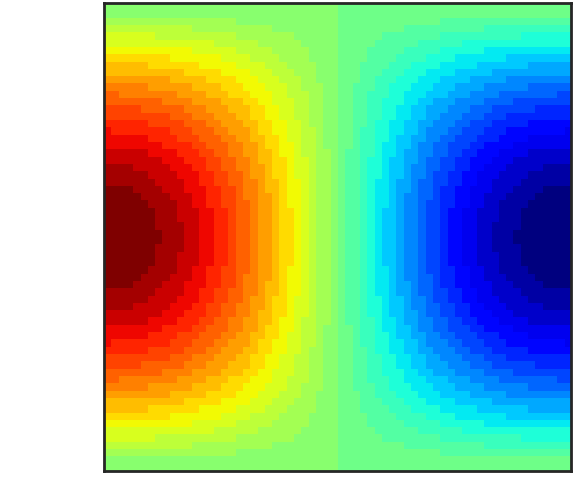}}
	\end{subfigure}%
	\begin{subfigure}[b]{.220\linewidth}
		\centering
		{\includegraphics[width=0.995\linewidth]{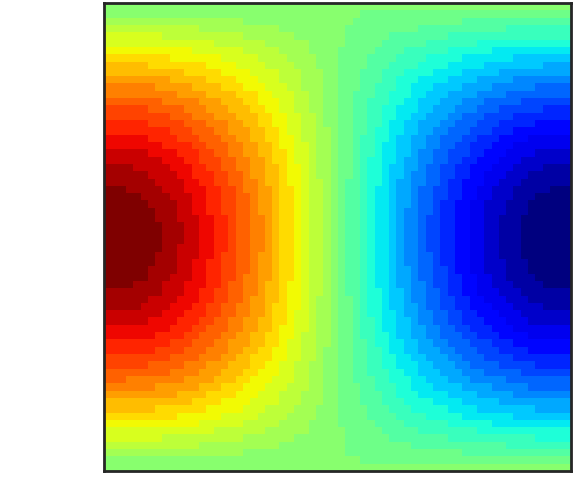}}
	\end{subfigure}%
    \begin{subfigure}[b]{.220\linewidth}
		\centering
		{\includegraphics[width=0.995\linewidth]{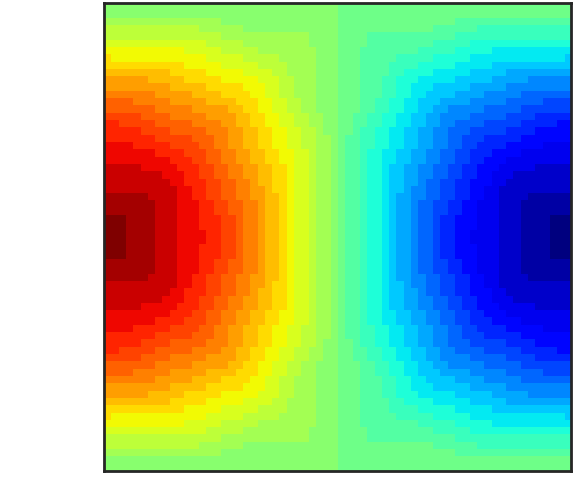}}
	\end{subfigure}\\
	\begin{subfigure}[b]{.220\linewidth}
		\centering
		{\includegraphics[width=0.995\linewidth]{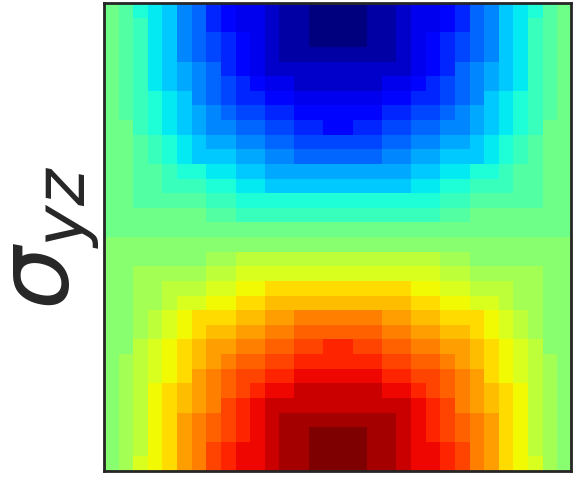}}
	\end{subfigure}%
	\begin{subfigure}[b]{.220\linewidth}
		\centering
		{\includegraphics[width=0.995\linewidth]{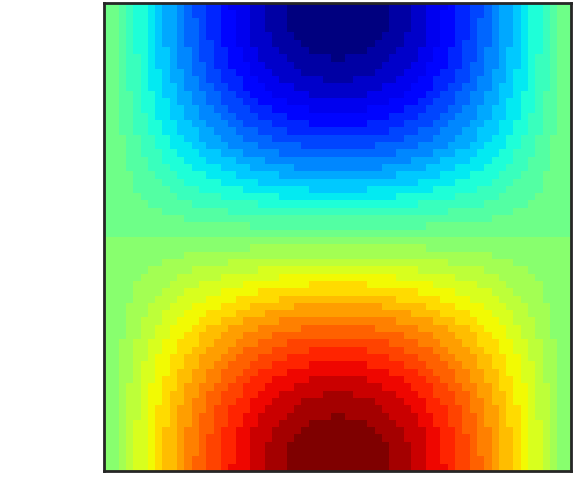}}
	\end{subfigure}%
	\begin{subfigure}[b]{.220\linewidth}
		\centering
		{\includegraphics[width=0.995\linewidth]{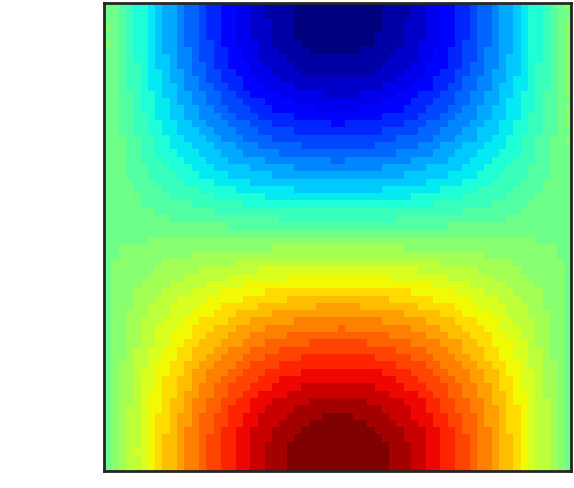}}
	\end{subfigure}%
    \begin{subfigure}[b]{.220\linewidth}
		\centering
		{\includegraphics[width=0.995\linewidth]{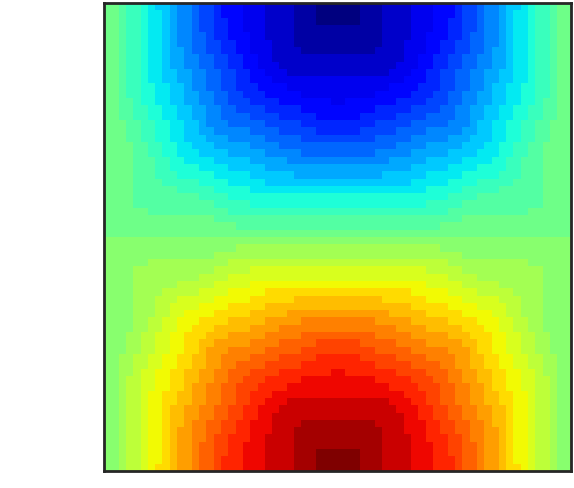}}
	\end{subfigure}\\
	\caption{
	The figure shows the snapshot of the elastodynamics' solution fields at a time step. 
	The LR input and HR Ground-truth are the coarse-scale and fine-scale solution fields obtained using the finite element method. 
	For a given LR input, the reconstructed solution from the framework is  highly accurate compared to the bi-linear interpolation method.
	}
	\label{fig:results}
\end{figure}

\subsection{Performance Evaluation}
This section focuses on the performance evaluation of the proposed framework. The low-resolution coarse-scale PDE solution generated in Section \ref{sec:input_data} is used as an input to the framework. In what follows, we compare the error measure $e$ for $t\ge0.02$ because the initial velocity condition is zero in this work. 

Figure \ref{fig:error_model} shows the error measure $e$ for all the HR state variables obtained from the framework. We note that the $\%$ error $e$ is less than $4\%$ for all the state variables at all times $t\le0.24$. We also note that the framework exhibits poor extrapolation capabilities as the errors become quite large after $t>0.24$. Figure \ref{fig:error_us} shows the error measure $e$ for the state variables when a simple bi-linear interpolation of the low-resolution data is performed to upscale the solution. We can see that the errors are relatively large for all of the state variables.

Figure \ref{fig:results} shows the super-resolved state variables obtained from the framework at a particular time $t=0.14$. We can notice that the super-resolved fields are indistinguishable from the HR ground truth data. At the same time, the simple bi-linear upscaling of the LR input data can be seen to significantly differ from the ground truth reference data. 

Therefore, we conclude that the proposed framework successfully enhanced the spatial and temporal resolution of the solution by a factor of $11$ and $2$, respectively.

\subsection{Speed-up}
Next, we calculate the speed-up obtained using the proposed framework to upsample the coarse-scale solution compared to obtaining the fine-scale solution using FEM. The FEM calculations are performed on a single-core AMD EPYC 7742 Processor. For the same simulation end time, the coarse-scale simulation ($\Delta x_c\approx0.176$ and $\Delta t_c = 0.005$) takes  $0.033$ seconds, whereas it takes $4.27$ seconds to run the fine-scale simulation ($\Delta x_f\approx0.0158$ and $\Delta t_f = 0.0025$). The above numbers are averaged over $10$ runs of the entire simulation. They do not include the time taken for mesh generation, node numbering, memory allocation, data I/O, or any other bookkeeping required by FEM.

On the other hand, after training, the inference time (averaged over $100$ inferences) for the spatial and temporal modules are $0.015$ and $0.017$ seconds, respectively. Therefore, using the proposed framework, it takes around $1.569$ seconds (including time taken for the coarse-scale simulation) to obtain the fine-scale PDE solution for the same simulation end time. Thus, we obtain a speed-up factor of around $2.72$ for the (relatively) simple test case discussed here. These inferences are performed on an NVIDIA Tesla V100-SXM2  GPU with $32$ GB RAM.

We emphasize here that the speed-up factor strongly depends on the complexity of the problem (linear vs. nonlinear) and the framework's spatial and temporal upscaling factors. 



\section{Conclusion}
\label{sec:conclusion}
In this work, we presented a novel unsupervised physics-informed machine learning framework, a first in the literature, that:
\begin{itemize}
\item enables spatial and temporal upscaling (resolution enhancement) of coarse-grained solutions to spatio-temporal PDEs while ensuring that the (upscaled) outputs satisfy the governing laws of the system.

\item easily allows imposition of any additional constraints (PDE or algebraic) in the framework.

\item is generalizable to non-rectangular domains by using  elliptic coordinate transformation as outlined in \cite{gao2020phygeonet}.

\item is amenable to scalability to clusters with multi-gpu nodes using \texttt{DistributedDataParallel} functionality in PyTorch   \cite{li2020pytorch} functionality for workloads that require substantial computational resources.

\end{itemize}

We demonstrated the framework's application to an elastodynamics problem under anti-plane strain conditions. The framework successfully enhanced the spatial and temporal resolutions of the coarse-scale input fields by a factor of $s=11$ and $k = 2$ in the space and time directions, respectively, while satisfying the physics-based constraints and yielding great accuracy (error $\le 4\%$). 

In the future, we aim to study the framework's application on a wide range of physical applications in fluid mechanics and compare the performance of the current framework with other super-resolution works (although supervised) in the literature \cite{ren2022physics, esmaeilzadeh2020meshfreeflownet}. Another interesting line of research to pursue would be the addition of ConvLSTM \cite{shi2015convolutional} in the temporal module to improve the  predicting capabilities of the framework beyond the convex hull of the traning set.




\section*{Research Data}
The low-resolution simulation data used in this work has been generated using Fenics \cite{alnaes2015fenics}. The source code for the proposed framework and the dataset used in this research can be found at \cite{arora_2022_STSR} upon acceptance of this paper.

\section*{Acknowledgment}
The work was conceptualized during the authors' time at Carnegie Mellon University (CMU). We gratefully acknowledge the Pittsburgh Supercomputing Center (PSC) for providing the computing resources that contributed to the research results reported in this paper.



\bibliographystyle{alpha} 
\bibliography{main} 

\end{document}